\pdfoutput=1
\documentclass[11pt]{article}

\usepackage[hyperref]{EMNLP2022}

\usepackage{times}
\usepackage{latexsym}
\usepackage{multirow}
\usepackage{adjustbox}
\usepackage{amssymb}
\usepackage{enumitem}
\usepackage{graphicx}
\usepackage{amsmath}
\usepackage{amsthm}
\usepackage{booktabs}

\usepackage[T1]{fontenc}
\usepackage[utf8]{inputenc}

\usepackage{microtype}

\usepackage{inconsolata}

\newcommand\blfootnote[1]{% 
\begingroup 
\renewcommand\thefootnote{}\footnote{#1}% 
\addtocounter{footnote}{-1}% 
\endgroup 
}

\title{PACIFIC: Towards Proactive Conversational Question Answering over Tabular and Textual Data in Finance \thanks{\hspace{1mm} This work is substantially supported by a grant from the Research Grant Council of the Hong Kong Special Administrative Region, China (Project Code: 14200719) and the Sea-NExT Joint Lab. Work done when Yang Deng was a visiting research scholar in the Sea-NExT Joint Lab.}}

\author{Yang Deng$^{1}$, Wenqiang Lei$^{2,\dagger}$, Wenxuan Zhang$^3$, Wai Lam$^1$, Tat-Seng Chua$^4$ \\
  $^1$The Chinese University of Hong Kong, $^2$Sichuan University,\\
  $^3$DAMO Academy, Alibaba Group, $^4$Sea-NExT Joint Lab, National University of Singapore \\
  \texttt{\{ydeng,wlam\}@se.cuhk.edu.hk}, \texttt{\{wenqianglei,isakzhang\}@gmail.com}
  \\}

\begin{document}
\maketitle
\begin{abstract}
To facilitate conversational question answering (CQA) over hybrid contexts in finance, we present a new dataset, named PACIFIC. 
Compared with existing CQA datasets, PACIFIC exhibits three key features: (i) proactivity, (ii) numerical reasoning, and (iii) hybrid context of tables and text. 
A new task is defined accordingly to study Proactive Conversational Question Answering (PCQA), which combines clarification question generation and CQA. 
In addition, we propose a novel method, namely UniPCQA, to adapt a hybrid format of input and output content in PCQA into the Seq2Seq problem, including the reformulation of the numerical reasoning process as code generation. 
UniPCQA performs multi-task learning over all sub-tasks in PCQA and incorporates a simple ensemble strategy to alleviate the error propagation issue in the multi-task learning by cross-validating top-$k$ sampled Seq2Seq outputs. 
We benchmark the PACIFIC dataset with extensive baselines and provide comprehensive evaluations on each sub-task of PCQA. 
\blfootnote{$^\dagger$ Corresponding author.}
\end{abstract}

\section{Introduction}
\begin{figure*}
\centering
\includegraphics[width=\textwidth]{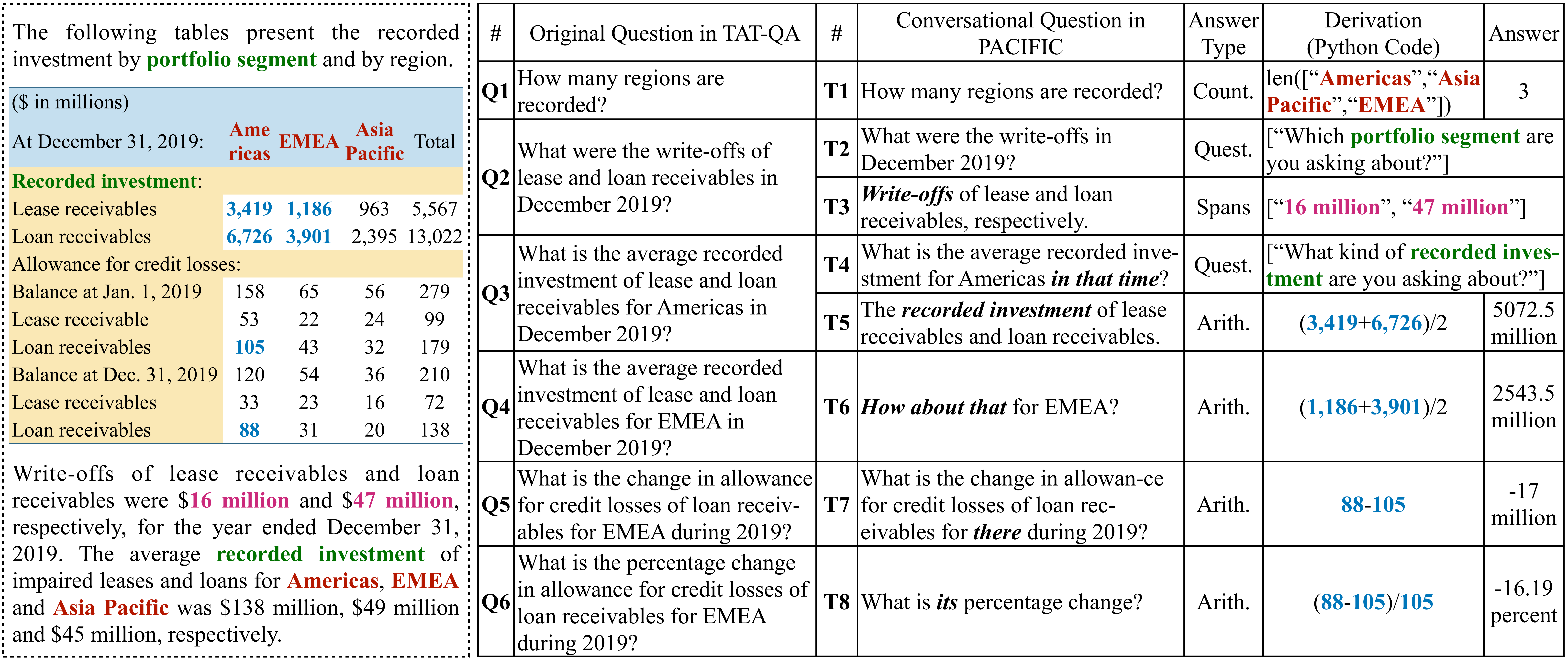}
\caption{An example of PACIFIC. The left dashed line box shows a hybrid context as the grounded document. The right solid line box shows the corresponding questions, responses with its answer types and derivation.}
\label{example}
\vspace{-0.3cm}
\end{figure*}

Financial question answering (QA) systems aim to answer user's instant queries by selecting appropriate information from financial documents, which often contain a hybrid of tabular and textual content, and performing complex quantitative analysis. 
Existing studies on financial QA~\cite{tat-qa,finqa,tat-dqa,acl22-tathqa} mainly focus on building \textit{single-turn QA} systems to \textbf{passively} respond to user queries. 
However, in real-world information-seeking applications~\cite{cis}, the system is expected to (i) answer highly context-dependent questions in a \textit{multi-turn conversation}, and (ii) \textbf{proactively} assist users in performing complicated information seeks. 
In an interactive setting, users tend to ask follow-up or co-referencing questions~\cite{acl20-followup-cqa,acl21-ask-convq} without repeating previous information, and provide a succinct or brief query that may be ambiguous or lack the necessary content. 
Especially in financial QA, the user queries often contain multiple constraints from different aspects for the concerned objective, as the examples shown in Fig.~\ref{example}. 
Even just missing one constraint may cause ambiguity. 
Therefore, a proactive conversational system that can help clarify the ambiguity is of great importance in financial QA. 

To this end, this paper introduces a new dataset to promote research into \textbf{P}ro\textbf{A}ctive \textbf{C}onversat\textbf{I}onal question answering in \textbf{FI}nan\textbf{C}e, named PACIFIC. PACIFIC is constructed by using the QA pairs in an expert-annotated financial QA dataset, TAT-QA~\cite{tat-qa}, as guidance to build conversation sessions with consecutive topics. As shown in Fig.~\ref{example}, we rewrite the original self-contained questions into conversational questions with anaphora (co-referencing among different turns) and ellipsis (omitting repeated words in the follow-up questions), as well as construct ambiguous questions that require clarification. 
Accordingly, we define a new task, named Proactive Conversational Question Answering (PCQA), which combines the problems of clarification question generation (CQG)~\cite{clariq} and conversational question answering (CQA)~\cite{coqa}. 
PCQA consists of three sub-tasks: (i) Given the user's query, the system first identifies whether the question is ambiguous (\textit{i.e.}, clarification need prediction). (ii) If so, the system will proactively ask a clarifying question to clarify the uncertainty (\textit{i.e.}, CQG). (iii) If not, it will directly answer the question (\textit{i.e.}, CQA).

Compared with existing datasets listed in Table~\ref{tab:dataset_comparison}, PACIFIC exhibits three key challenges: (i) \textit{proactivity}: the system needs to proactively assist the user to clarify their question intent by asking clarifying questions; (ii) \textit{numerical reasoning}: there are a large number of questions that require numerical reasoning to answer; and (iii) \textit{hybrid context}: the grounded document is composed by both tabular and textual content. 

To tackle these challenges, we propose a novel method, named UniPCQA, to unify all sub-tasks in PCQA as a sequence-to-sequence (Seq2Seq) problem. 
Specifically, we reformulate the numerical reasoning process in financial question answering as a code generation task, which captures the input knowledge (\textit{e.g.}, figures or entities) and condenses their numerical reasoning relations (\textit{e.g.}, arithmetic operators) into a piece of executable code (\textit{e.g.}, \texttt{Python}). 
We further design specific input and output representations to adapt a hybrid of tabular, textual, and arithmetic content into the Seq2Seq framework. 
In addition, UniPCQA can perform multi-task learning over all sub-tasks to enable the proactive detection of the need for clarification. 
Finally, we propose an ensemble strategy, named Consensus Voting, to alleviate the error propagation issue in the multi-task learning by cross-validating the top-$k$ sampled Seq2Seq outputs. The main contributions of this paper are:
\begin{itemize}[leftmargin=*]
    \item To study the proactivity in financial question answering, we propose a novel dataset, namely PACIFIC, for conversational question answering over tabular and textual contexts, and define the problem of PCQA.  
    \item We reformulate the numerical reasoning process as code generation and propose a unified hybrid Seq2Seq framework, namely UniPCQA, to handle the hybrid contexts and diverse responses in PCQA. 
    \item We benchmark the PACIFIC dataset with extensive baselines and provide comprehensive evaluations on each sub-task of PCQA. Despite the effectiveness of UniPCQA, the performance is far behind human experts, showing that PACIFIC presents a challenging problem for future studies.  
\end{itemize}

\section{Related Works}

\begin{table}[]
\setlength{\abovecaptionskip}{5pt}   
\setlength{\belowcaptionskip}{0pt}
    \centering
    \begin{adjustbox}{max width=0.48\textwidth}
    \begin{tabular}{llllll}
    \toprule
        Dataset & Domain &  Turn & Modality & Proact. & NR \\
        \midrule
         Hybrid-QA &  General & Single & Table/Text & $\times$ & $\times$  \\
         OTT-QA &General & Single&Table/Text&$\times$&$\times$\\
         FinQA&Finance& Single&Table/Text&$\times$&$\checkmark$\\
         TAT-QA&Finance& Single&Table/Text&$\times$&$\checkmark$\\
         \midrule
         SQA&General& Multi& Table&$\times$&$\checkmark$\\
         QuAC&General& Multi&Text&$\times$&$\times$\\
         CoQA&General& Multi& Text&$\times$&$\times$\\
         Abg-CoQA&General& Multi& Text&$\checkmark$&$\times$\\
         HybriDial.&General& Multi&Table/Text&$\times$&$\times$\\
         MMConvQA&General& Multi&Table/Text/Image&$\times$&$\times$\\
         ConvMix&General& Multi&Table/Text/KB&$\times$&$\times$\\
         \midrule
         \textbf{PACIFIC}&Finance& Multi&Table/Text&$\checkmark$&$\checkmark$\\
         \bottomrule
    \end{tabular}
    \end{adjustbox}
    \caption{Comparison of PACIFIC and related QA and CQA datasets. ``NR'' denotes Numerical Reasoning. }
    \label{tab:dataset_comparison}
\vspace{-0.4cm}
\end{table}

\noindent\textbf{Conversational Question Answering} ~
Evolving from single-turn QA tasks~\cite{hybridqa,tois21-ydeng}, CQA aims at interactively answering multiple turns of information-seeking questions according to the given document~\cite{coqa,quac}. 
Common challenges in CQA include the anaphora and  ellipsis issue~\cite{seqqa,acl20-followup-cqa,acl21-ask-convq}. 
To this end, several attempts have been made on developing end-to-end CQA models with dialogue history tracking~\cite{sigir19-coqa-history,aaai21-coqa-history}. 
Another group of works emphasizes the importance of query rewriting in CQA~\cite{wsdm21-query-cqa,ecir22-query-cqa,naacl21-query-cqa,acl21-coqa-dependency}, which generates self-contained questions for performing single-turn QA. 
In addition, beyond simply focusing on one kind of information source, it has received increasing attentions to investigate CQA over hetergeneous sources~\cite{mmconvqa,convmix}.

\noindent\textbf{Proactive Conversational Systems} ~
Early studies on conversational systems basically develop dialogue systems that passively respond to user queries, including all the CQA studies discussed above. 
As for conversational recommendation~\cite{ear,scpr,sigir21-unicorn} and goal-oriented dialogues~\cite{sigir22-poactive,mg-crs}, policy learning or goal planning attaches great importance in building a proactive conversational system for promptly adjusting dialogue strategies or soliciting user intents. 
Recently, many efforts have been made on CQA systems that can proactively assist users to clarify the ambiguity or uncertainty in their queries by asking clarifying questions~\cite{cikm21-cqg,www2020-clari,ictir21-qgpt,ecir22-clari}.   
Several datasets such as ClariQ~\cite{clariq} and Abg-CoQA~\cite{akbc} have been constructed to facilitate this line of research. 
However, these datasets solely target at the clarification question generation (CQG) or clarification-based CQA problem. 
To stimulate progress of building the whole system for proactive CQA, we define the PCQA task, which unifies CQG and CQA.

\noindent\textbf{Numerical Reasoning} ~
Numerical reasoning is the key to many NLP applications~\cite{naacl21-nr-survey,emnlp21-findings-nr}, especially in QA, such as Mathematical QA~\cite{drop,mathqa} and Financial QA~\cite{tat-qa,finqa}.
Early works typically design specialized operation or reasoning modules for handling different types of questions~\cite{emnlp19-numerical,emnlp19-numerical2,emnlp19-numnet}.  Despite the effectiveness, it is challenging for them to scale to different numerical reasoning scenarios due to their task-specific designs. 
Recent years have witnessed many advanced approaches to injecting the numerical reasoning skills into pre-trained language models (PLMs), by post-training~\cite{acl20-nr-lm,program-execute} or prompt-based learning~\cite{chain-of-thought,self-consistency}. 
However, these methods are developed to perform numerical reasoning over texts. 
\newcite{acl21-table2text} investigate template-based table representations for numerical reasoning in PLMs-based table-to-text generation. 
In this paper, we propose to handle numerical reasoning as the code generation task over hybrid contexts.

\section{PACIFIC Dataset Creation}
\subsection{Annotation \& Quality Control}\label{sec:dataset}
Similar to the dataset creation process of other CIS datasets, such as HybriDialogue~\cite{hybridialogue} from OTT-QA~\cite{ott-qa} and MMConvQA~\cite{mmconvqa} from MMQA~\cite{mmqa}, we build the PACIFIC dataset from the TAT-QA dataset by using its question-answer pairs as guidance for constructing conversation sessions. 
There are on average 6 individual question-answer pairs shared with the same grounded contexts in TAT-QA, which are integrally regarded as one conversation session. 
However, we construct the conversation session in a different way from the traditional manner where a complex single-turn question is decomposed into multiple context-dependent simple questions~\cite{hybridialogue,mmconvqa}, since this manner may discard the nature of financial QA. 
Instead, we rewrite each question into one conversational question, which not only increases the efficiency of dataset construction, but also preserves the quality and difficulty of the dataset with expert-annotated answers and informative user queries. 

Due to the space limitation, the overall pipeline for PACIFIC creation is presented in Appendix~\ref{app:dataset}. An example is presented in Figure~\ref{example} with its original sample in TAT-QA. 
For each conversation sample, two annotators are asked to build a natural and consecutive conversation session. They are well-educated postgraduate students majored in finance or similar disciplines. 
The first annotator serves as the seeker to perform the annotation tasks; while the second annotator plays the role of the agent to provide clarifying questions. 
The instructions given to the first annotator are as follows:

\begin{table}
    \centering
    \begin{adjustbox}{max width=0.48\textwidth}
    \begin{tabular}{lrrr}
    \toprule
        PACIFIC/TAT-QA & Train & Dev & Test \\
        \midrule
         \# Dialogues& 2,201/-& 278/- & 278/-\\
         \# Turns (QA pairs)&15,087/13,215&1,982/1,668&1,939/1,669\\
         \# Clarifying turns&1,872/-&320/-&270/-\\
         Avg. turns / dialogue&6.9/-&7.1/-&7.0/-\\
         Avg. words / question&9.6/12.5&9.0/12.4&9.4/12.4\\
         Avg. words / answer&4.6/4.1&4.6/4.1&4.8/4.3\\
         \bottomrule
    \end{tabular}
    \end{adjustbox}
    \caption{Data statistics of PACIFIC.}
    \label{tab:dataset}
\vspace{-0.3cm}
\end{table}

\begin{table}
\fontsize{8}{9.5}\selectfont
    \centering
    \begin{adjustbox}{max width=0.48\textwidth}
    \begin{tabular}{lrrrr}
    \toprule
         & Table & Text & Table-text & Total \\
         \midrule
       Span  & 1,797 & 3,497 & 1,842 & 7,136\\
       Spans & 777 & 258 & 1,037 & 2,072\\
       Counting & 106 & 5&266 & 377\\
       Arithmetic & 4,744 & 143 & 2,074 & 6,961\\
       Question & 1,293&270&899&2,462\\
       Total & 8,717 &4,173&6,118&19,008\\
       \bottomrule
    \end{tabular}
    \end{adjustbox}
    \caption{Number of questions regarding different answer types and sources in PACIFIC.}
    \label{tab:question_distribution}
\vspace{-0.3cm}
\end{table}

1) \textit{Organize Conversation Sessions}. Given the same hybrid context, set up a conversation session with consecutive topics from multiple individual QA pairs. Two questions that share the same entities
are regarded as talking about the same topic. For example, \textbf{Q1}, \textbf{Q2}, and \textbf{Q3} are concerned about the same time (\textit{December 2019}), while \textbf{Q4}, \textbf{Q5}, and \textbf{Q6} are asking about the same region (\textit{EMEA}). We, thus, order these questions into adjacent turns. 

2) \textit{Rewrite Conversational Questions}. If consecutive questions share the same entities, rewrite the original self-contained questions to produce conversational questions with anaphora and ellipsis. For example, the only difference between \textbf{Q3} and \textbf{Q4} is the concerned region (\textit{Americas} \& \textit{EMEA}). After the rewriting, \textbf{T6} becomes ``\textit{How about that for EMEA?}" without the repeated content in \textbf{T4}.

3) \textit{Construct Ambiguous Questions}. If the question contains multiple entities, rewrite it to construct an ambiguous question by omitting one of the entities that can introduce ambiguity. For example, \textbf{Q3} is asking about the average value under multiple constraints. In \textbf{T4}, the portfolio segment (\textit{Lease and loan receivables}) is omitted to construct an ambiguous question that required clarification. 

Given the set of reconstructed questions, the second annotator is served as the agent to \textit{Provide Clarification Questions}: \textit{i.e.}, ask a clarification question in terms of the omitted entity. Subsequently, the omitted entity will be the seeker's query in the next turn, as \textbf{T3} and \textbf{T5} in Fig.~\ref{example}.

To ensure the quality of annotation in PACIFIC, we ask two verifiers to validate each turn in the constructed conversations. If any mistake or problem is found, \textit{e.g.}, the constructed conversation is incoherent, the annotator will be asked to fix it until the annotation passes the checks by the two verifiers. The first-round validation captures 212 mistakes (212/19,008=1.1\%), and the inter-annotator agreement between the two verifiers is 0.62.

\subsection{Statistical Analysis}
Finally, we obtain a total of 2,757 conversations over the hybrid contexts, which contains 19,008 corresponding QA pairs in total and an average of 7 turns of QA in each conversation. The train-dev-test split is the same as TAT-QA. We present the data statistics of PACIFIC in Table~\ref{tab:dataset}, and the question distribution regarding different answer types and sources in Table~\ref{tab:question_distribution}. 
Compared with TAT-QA, PACIFIC contains 2,462 more QA turns for asking clarification questions (2,462/19,008=13.0\%)\footnote{The proportion of clarification interactions is close to existing CQG datasets, \textit{e.g.}, 16.1\% in \cite{clariq} and 11.5\% in \citet{akbc}}.
The average length of the questions in PACIFIC is shorter than that in TAT-QA, which means that the conversational questions are more succinct and brief. Conversely, the average length of the answers in PACIFIC is longer than that in TAT-QA, due to the incorporation of clarification questions.

\subsection{Problem Definition}
\begin{figure}
\centering
\includegraphics[width=0.5\textwidth]{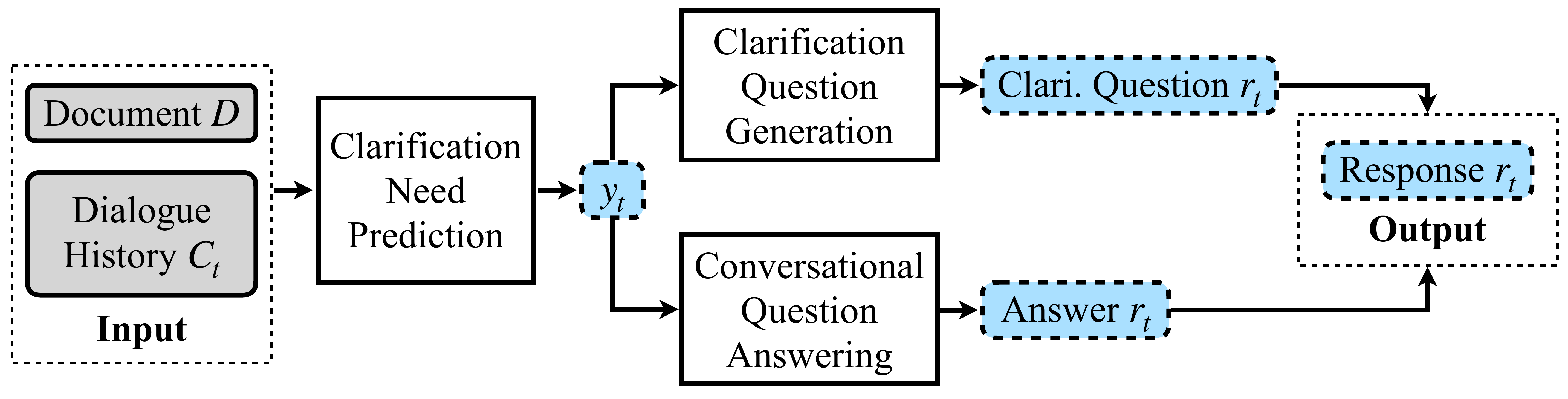}
\caption{Proactive conversational question answering.}
\label{problem}
\vspace{-0.3cm}
\end{figure}

We introduce the \textbf{Proactive Conversational Question Answering (PCQA)} task, which unifies two tasks: (I) Clarification Question Generation and (II) Conversational Question Answering. 
Given the conversation history $C_t=\{q_1,r_1,...,q_{t}\}$ and the grounded document $D=\{E,T\}$ consisting of both textual contexts $E$ and structured table $T$, the goal is to generate the response $r_t$ at the current turn $t$. 
As shown in Fig.~\ref{problem}, the overall task can be decomposed into three sub-tasks:

    1) \textit{Clarification Need Prediction} (CNP) aims to predict the binary label $y$ to determine whether to ask a question for clarifying the uncertainty. Otherwise the query $q_t$ can be directly responded to. 
    
    2) \textit{Clarification Question Generation} (CQG) will generate a clarification question as the response $r_t$, if CNP detects the need for clarification. 
    
    3) \textit{Conversational Question Answering} (CQA) will directly produce the answer as the response $r_t$, if it is not required for clarification.

It is worth noting that the PACIFIC dataset can be adopted for the evaluation of both end-to-end and pipeline-based PCQA methods, as well as the separated evaluation on each sub-task.

\section{Method}
We introduce the UniPCQA model, which unifies all sub-tasks in PCQA as the Seq2Seq problem and performs multi-task learning among them. 

\subsection{Numerical Reasoning as Code Generation}
One of the key challenges of PACIFIC is the requirement to conduct numerical reasoning, due to the large proportion of questions involving numerical calculation. 
However, existing methods proposed for financial question answering suffer from two main issues: 1) they rely heavily on hand-crafted designs for numerical operators~\cite{tat-qa} or symbolic programs~\cite{finqa}, which are hard to be generalized to complex numerical calculation; 2) the knowledge from large-scale PLMs cannot be fully utilized for the down-stream problem of numerical reasoning, due to the large gap between them. 

In the light of these issues, we formulate the numerical reasoning process as the code generation task, which aims to capture the input knowledge (\textit{e.g.}, figures or entities) and condense their numerical reasoning relations (\textit{e.g.}, arithmetic operators) into a piece of executable code. 
Take \texttt{Python} as an example. 
\texttt{Python} can handle the derivation with different kinds of operations, such as arithmetic, counting, enumeration, etc.  
The addition, subtraction, multiplication and division operators are denoted by +, -, *, and /, respectively. The \texttt{len()} function that returns the number of items in an object can be used for the counting operation. 
To be consistent, we also regard span-based and question-based responses as a \texttt{list()} of items in \texttt{Python} for code generation.  
Examples of \texttt{Python} code for different types of answers are shown in Fig.~\ref{example}. 

Without the need for designing another execution algorithm~\cite{tat-qa,finqa}, the generated \texttt{Python} code can be directly executed by the \texttt{eval()} function to derive the final answer $r_t$, as the following examples: 
\begin{align*}
    &\texttt{eval}((36.6-20.5)/20.5) \rightarrow 0.7854\\
    &\texttt{eval}(\texttt{len}([``2018",``2019"])) \rightarrow 2
\end{align*}
Therefore, we can reconstruct the target \texttt{Python} code from the original answer derivation, according to the \texttt{Python} syntax, which can be easily generalized into different types of numerical calculation. The numerical reasoning process can not only get free of manually designed operators or programs, but also leverage the knowledge from PLMs, especially those code-related PLMs, such as CodeT5~\cite{codet5}.

\begin{figure}
\centering
\includegraphics[width=0.5\textwidth]{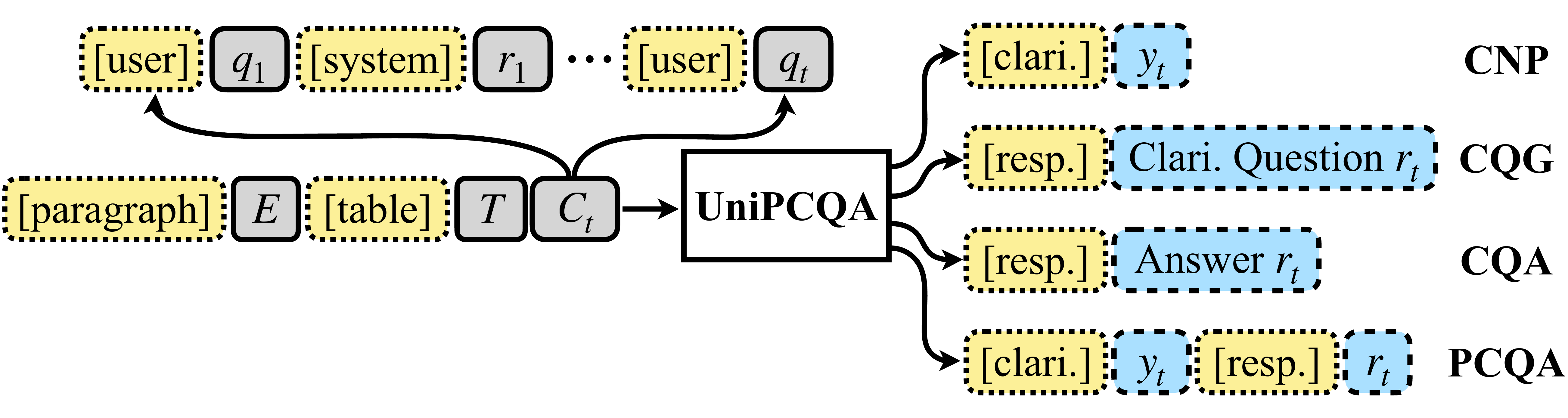}
\caption{Overview of the input/output for UniPCQA. Note that the first three outputs denote the output in the single-task learning setting for each sub-task, while the last one denotes that in the multi-task learning setting.}
\label{method}
\vspace{-0.4cm}
\end{figure}

\subsection{Hybrid Seq2Seq Generation}
In financial PCQA, the input sequence contains both textual and tabular content, while the output sequence can be a piece of code, a natural language sentence, or even a mix of code and text. In order to handle all sub-tasks in PCQA, we design the hybrid input/output representations for a unified Seq2Seq framework. Specifically, we add special tokens to indicate different types of information as well as specify each sub-task. Assuming that the grounded textual context is $E=\{p_1,...p_k\}$ and the grounded structure table is $T=\{c_{11},...,c_{1n},...,c_{m1},...,c_{mn}\}$, then the input can be linearized as follows: 
\begin{align*}
    ``&\texttt{[paragraph]}~ p_1~ \texttt{</p>}~ ...~\texttt{</p>} ~ p_k ~\texttt{</p>} ~\texttt{[table]} \\
    &c_{11} : c_{12}~ |~ ... ~| ~c_{1n} ~\texttt{</t>}~ ...~ c_{m1} : c_{m2} ~|~ ... ~|~ c_{mn}\\
    & \texttt{[user]}~ q_1 ~\texttt{[system]}~ r_1~ ... ~\texttt{[user]}~ q_t "
\end{align*}
As shown in Fig.~\ref{method}, this Seq2Seq formulation can be applied to each sub-task, or perform multi-task learning of all sub-tasks in order. The output sequence for multi-task learning is represented as:
\begin{equation*}
    ``\texttt{[clari.]} ~ y ~ \texttt{[resp.]} ~ r_t"
\end{equation*}
where $y\in\{\texttt{True},\texttt{False}\}$, and $r_t$ will be a clarification question or a piece of code accordingly. 

UniPCQA can be initialized with weights from any generative PLM, \textit{e.g.}, T5~\cite{t5}. 
Given a training sample $(C_t,D,o)$, the model is trained to maximize the sequential log-likelihood:
\begin{equation}
    \mathcal{L}_\theta  = \sum\nolimits^{L}_{l=1}\log p_\theta(o_{l}|o_{<l};C_t,D),
\end{equation}
where $\theta$ denote the model parameters, $L$ is the maximum target sequence length, and $o$ is the target sequence according to the target task.

\subsection{Consensus Voting}
As UniPCQA solves the end task using multi-task learning in sequential order, the error in the previous task may be propagated to the latter one. Specifically, if the model makes a wrong prediction in the CNP task, the model will generate an inappropriate response at the end.

Inspired by the Self-Consistency strategy~\cite{self-consistency} for improving the few-shot learning accuracy of PLMs, we investigate a similar ensemble-based strategy, namely Consensus Voting, to alleviate the error propagation issue in the multi-task learning. 
Specifically, Consensus Voting samples a set of candidate sequences $O=\{o_i: i\in 1,...,N$\} generated by the PLM, which contains a diverse set of multi-task results as well as different reasoning paths, instead of using Greedy Decode. 
We then select the final response by ensembling the derived responses from $O$ based on plurality voting:
\begin{equation}
    r_t=\arg\max\nolimits_{o_i\in O}\sum\nolimits^N_{j=1}\mathbb{I}(\sigma(o_j)=\sigma(o_i)),
\end{equation} 
where $\sigma(\cdot)$ denotes the execution of deriving the answer from the generated sequence, \textit{e.g.}, \texttt{eval()}. 

The motivation is that it will be difficult for the sampled outputs to reach a consensus if the user query is ambiguous, since the decoder will be confused about how to generate a correct derivation with incomplete information. 
At this time, the plurality vote will tend to ask a clarification question. 
In addition, the same answer can be obtained by executing different derivations in some cases. As shown in Fig.~\ref{example}, different extraction orders of three regions can lead to the same answer in T1, \textit{e.g.}, $\texttt{len}\text{([``Americas", ``EMEA", ``Asia Pacific"])} =  \texttt{len}\text{([``EMEA", ``Americas", ``Asia Pacific"])}$. So does the derivation in T8, \textit{i.e.},  $(88-105)/105 = 88/105-1$. Therefore, if there are multiple generated derivations that lead to the same answer, this answer will get higher votes.

\section{Experiments}
We first evaluate methods on two widely-studied tasks in conversational information seeking, including (I) clarification question generation (CQG) and (II) conversational question answering (CQA). 
Then we benchmark the overall performance of proactive conversational question answering (PCQA) on PACIFIC. 

\subsection{Implementation}
We evaluate UniPCQA with T5$_\text{base}$ as the baseline. To study the effectiveness of the reformulation of code generation, we further adopt CodeT5$_\text{base}$~\cite{codet5} for evaluation, which is a unified encoder-decoder model pre-trained with both code-related understanding and generation tasks. 
Following previous studies~\cite{top-k-sample,iclr20-top-k}, we apply top-$k$ sampling with temperature $T=0.5$ and $k=40$ to sample a diverse set of decoded sequences. 
For Consensus Voting, we sample $N=40$ outputs, while the baseline is to apply Greedy Decode to generate a single output. 
More implementation details can be found in Appendix~\ref{app:implement}. 

\subsection{Task I: Clarification Question Generation}
The CQG task is commonly performed in two steps: 1) clarification need prediction (CNP), and 2) clarification question generation. 

\subsubsection{Baselines and Evaluation Metrics}
Following ClariQ~\cite{clariq}, a popular CQG challenge, we include BERT$_\text{large}$~\cite{bert} and RoBERTa$_\text{large}$~\cite{roberta} based classifiers as baselines, and use Precision, Recall, and F1 for CNP evaluation. 
For CQG, we compare to several CQG baselines in latest studies, including  Template-based Question Generation (TB)~\cite{www2020-clari}, CopyTrans.~\cite{cikm21-cqg}, and Q-GPT~\cite{ictir21-qgpt}, and adopt ROUGE-2 (F1), Exact Match (EM), and token-level F1 as evaluation metrics.  
Note that we simply flatten the table into a sequence by row followed by tokens from the paragraphs for all the baselines, which is also applied to the baselines in the following evaluation. More details about baselines can be found in Appendix~\ref{app:baseline}. 

\begin{table}
\fontsize{8}{9.5}\selectfont
    \centering
    \begin{adjustbox}{max width=0.48\textwidth}
    \begin{tabular}{lcccccc}
    \toprule
       \multirow{2}{*}{Method}  & \multicolumn{3}{c}{Dev} & \multicolumn{3}{c}{Test}\\
       \cmidrule(lr){2-4}\cmidrule(lr){5-7}
        & P & R & F1 & P & R & F1\\
        \midrule
        BERT$_\text{large}$ & 84.5 & 85.4&84.9& 80.0&83.9&81.7\\
        RoBERTa$_\text{large}$&\underline{93.1}&\underline{89.4}&\underline{91.2}&\underline{90.0}&\underline{90.8}&\underline{90.2}\\
        \midrule
        UniPCQA (T5)&\textbf{93.7}&\textbf{91.0}&\textbf{92.3}&\textbf{90.6}&\textbf{91.6}&\textbf{91.1}\\
        \bottomrule
    \end{tabular}
    \end{adjustbox}
    \caption{Results on Clarification Need Prediction.}
    \label{tab:cnp}
\vspace{-0.3cm}
\end{table}

\begin{table}
    \centering
    \begin{adjustbox}{max width=0.48\textwidth}
    \begin{tabular}{lcccccc}
    \toprule
       \multirow{2}{*}{Method}  & \multicolumn{3}{c}{Dev} & \multicolumn{3}{c}{Test}\\
       \cmidrule(lr){2-4}\cmidrule(lr){5-7}
        & ROUGE & EM & F1 & ROUGE & EM & F1\\
        \midrule
        BERT+TB&69.8&36.3&75.4&67.8&33.2&72.8\\
        CopyTrans.&70.3&39.4&75.4&68.1&37.9&73.2\\
        Q-GPT&\underline{86.5}&\underline{67.8}&\underline{90.5}&\underline{83.9}&\underline{63.4}&\underline{87.8}\\
        \midrule
        UniPCQA (T5)&\textbf{90.7}&\textbf{76.9}&\textbf{93.4}&\textbf{87.8}&\textbf{71.1}&\textbf{91.1}\\
        \bottomrule
    \end{tabular}
    \end{adjustbox}
    \caption{Results on Clarification Question Generation.}
    \label{tab:cqg}
%x\vspace{-0.3cm}
\end{table}

\subsubsection{Experimental Results}
Table~\ref{tab:cnp} presents the experimental results on the CNP task, showing that a stronger PLM leads to better performance in this binary classification task. 

Table~\ref{tab:cqg} summarizes the experimental results on the CQG task. 
Baseline methods can achieve relatively higher scores for ROUGE and F1, due to the similar expressions among different clarification questions. However, without using PLMs, CopyTrans. has a similar performance as the template-based method (BERT+TB). 
UniPCQA outperforms Q-GPT by a noticeable margin, indicating the effectiveness of the hybrid input sequence construction in such an CQG task based on hybrid contexts.

\subsection{Task II: Conversational QA}
Following previous  studies~\cite{wsdm21-query-cqa,acl21-coqa-dependency}, we compare to both end-to-end and pipeline-based methods. End-to-end methods adopt a single QA or CQA model to encode the document and the whole conversation history, while pipeline-based methods decompose the CQA task into Query Rewriting (QR) and single-turn QA that are solved by different models.

\begin{table}
    \centering
    \begin{adjustbox}{max width=0.48\textwidth}
    \begin{tabular}{llcccc}
    \toprule
       \multirow{2}{*}{QR Model}  & \multirow{2}{*}{QA Model} & \multicolumn{2}{c}{Dev} & \multicolumn{2}{c}{Test}\\
       \cmidrule(lr){3-4}\cmidrule(lr){5-6}
         & & EM & F1 & EM & F1\\
        \midrule
        \multirow{5}{*}{Gold}
        &NumNet+ V2 &38.1&48.3&37.0&46.9\\
        &\textsc{TagOp} & 55.2 & 62.7 & 50.1 & 58.0 \\
        &TaCube&57.7&66.2&-&-\\
        &\textsc{PoEt-SQL}&59.1&65.9&-&-\\
        &UniPCQA (T5)&65.3 & 72.9 & 62.3 & 71.1\\
        &UniPCQA (CodeT5)&\textbf{68.2}&\textbf{75.5}&\textbf{63.9}&\textbf{72.2}\\
        
        \midrule
        \midrule
        Original & \multirow{4}{*}{\textsc{TagOp}} & 39.4& 46.6& 34.7 &43.2 \\
        Trans.++&& 41.8&48.1&36.2&43.9\\
        T5&&42.0&48.4&36.6&44.2\\
        T5$^*$&&50.0&56.6&46.2&54.2\\
        \midrule
        \multirow{3}{*}{End-to-end} &NumNet+ V2 &30.2&39.0&27.7&36.9\\
        & \textsc{TagOp} &45.6&53.2&43.3&50.4\\
        & HAE (BERT$_\text{large}$) &20.3&30.6&18.2&25.4\\
        \midrule
        \midrule
        \multirow{2}{*}{End-to-end} &UniPCQA (T5)  & 62.6&69.7&58.9&67.3\\
        &UniPCQA (CodeT5) &\textbf{64.7}&\textbf{72.0}&\textbf{59.8}&\textbf{67.9}\\
        \bottomrule
    \end{tabular}
    \end{adjustbox}
    \caption{Results on Conversational QA. $^*$ denotes that the QR model is trained on the QR data from PACIFIC.}
    \label{tab:cqa}
\vspace{-0.3cm}
\end{table}

\subsubsection{Baselines and Evaluation Metrics}

We adopt the following QR methods for comparisons: Original, Trans.++~\cite{wsdm21-query-cqa}, and T5~\cite{cqr-t5,acl21-coqa-dependency}. 
QR methods are trained on the QReCC dataset~\cite{naacl21-query-cqa}.  
We include three QA/CQA models for comparisons: HAE~\cite{sigir19-coqa-history}, 
NumNet+ V2~\cite{emnlp19-numnet}, and \textsc{TagOp}~\cite{tat-qa}. 
Details can be found in Appendix~\ref{app:baseline}. 
 
In addition, we report the performance of using ground-truth self-contained questions (\textbf{Gold}) as input for single-turn QA models. This is equivalent to their performance on the TAT-QA dataset, including two latest results, \textit{i.e.}, TaCube~\cite{tacube} and \textsc{PoEt-SQL}~\cite{program-execute}. 

Following previous studies on financial question answering~\cite{tat-qa}, we use EM and numeracy-focused F1 score~\cite{drop} for the CQA evaluation.

\subsubsection{Experimental Results}

The CQA results are summarized in Table~\ref{tab:cqa}. There are several noticeable observations: 

(1) A good QR model can lead to better performance on the CQA task for pipeline-based methods, where using ground-truth self-contained questions (Gold) can be regarded as an estimate of the upper bound for these methods. 
For a fair comparison, the QR models should not be trained on the QR data from PACIFIC. Therefore, pipeline-based methods barely work on PACIFIC when using an out-of-domain QR model. We also report the performance of each QR method in Appendix~\ref{app:qr}.  

(2) Conventional CQA methods, \textit{e.g.}, HAE, fail to achieve promising results on PACIFIC, due to the inability of handling numerical reasoning.

(3) UniPCQA not only achieves the best performance on the original TAT-QA dataset, but also outperforms both pipeline-based and end-to-end methods on the CQA task, \textit{i.e.}, PACIFIC. 
These results show the superiority of UniPCQA in handling both single-turn and conversational finance QA problems.

\begin{table}
    \centering
    \begin{adjustbox}{max width=0.48\textwidth}
    \begin{tabular}{lcccc}
    \toprule
       \multirow{2}{*}{Method}  & \multicolumn{2}{c}{Dev} & \multicolumn{2}{c}{Test}\\
       \cmidrule(lr){2-3}\cmidrule(lr){4-5}
         &  EM & F1 & EM & F1\\
        \midrule
        DialoGPT&25.2&32.1&22.6&30.7\\
        FinQANet&40.3&47.2&38.0&45.5\\
        T5+\textsc{TagOp}&\underline{49.6}&\underline{56.1}&\underline{46.6}&\underline{52.3}\\
        \midrule
        UniPCQA (T5)&60.0 & 68.1 &56.9&65.4\\
        UniPCQA (T5) + MTL&61.6&70.3&58.7&67.5\\
        UniPCQA (T5) + MTL + CV&62.1&70.7&59.4&68.3\\
        \midrule
        UniPCQA (CodeT5)&63.2&71.4&60.4&68.5\\
        UniPCQA (CodeT5) + MTL&64.0&72.1&61.0&69.4\\
        UniPCQA (CodeT5) + MTL + CV&\textbf{64.5}&\textbf{72.6}&\textbf{61.9}&\textbf{70.2}\\
        \midrule
        Human&-&-&86.3&92.1\\
        \bottomrule
    \end{tabular}
    \end{adjustbox}
    \caption{Overall evaluation on PCQA.}
    \label{tab:pcqa}
    \vspace{-0.3cm}
\end{table}

\subsection{Overall Evaluation on PCQA}
\subsubsection{Baselines and Evaluation Metrics}
Since this is a preliminary attempt on PCQA over hybrid contexts, we implement several alternative solutions for method comparisons, including two end-to-end generation methods, \textbf{DialoGPT}~\cite{hybridialogue} and  \textbf{FinQANet}~\cite{finqa}, as well as one pipeline-based method, \textbf{T5+\textsc{TagOp}}~\cite{tat-qa}. 
As a union of CQG and CQA, we adopt their shared evaluation metrics for the evaluation of PCQA models, including EM and numeracy-focused token-level F1. More details about baselines can be found in Appendix~\ref{app:baseline}. 

\subsubsection{Experimental Results}
Table~\ref{tab:pcqa} presents the experimental results on the PCQA task. Among the baselines, due to the inability of performing numerical reasoning, DialoGPT performs much worse than FinQANet and T5+\textsc{TagOp}.  T5+\textsc{TagOp} achieves a better performance than FinQANet, as its operators are specifically designed for the TAT-QA dataset, which can also be effectively applied to PACIFIC. Finally, UniPCQA substantially outperforms all these baselines, \textit{i.e.}, 56.9/65.4 vs. 46.6/52.3 in EM/F1. 
In addition, UniPCQA is more flexible to different numerical calculations without the reliance on manually designed operators or programs and additional algorithms for executing the system outputs. 

Among different variants of UniPCQA, CodeT5 achieves a better performance than T5 with the same size of model parameters, which indicates that UniPCQA effectively leverages the knowledge from the code-related pre-training tasks. 
The multi-task learning (MTL) improves the performance by explicitly learning from the clarification need labels. Further, the error propagation issue introduced by the MTL is alleviated by the Consensus Voting strategy (CV). However, compared with the performance of human experts (Human), there is still much room for improvement.

\begin{figure}
\centering
\includegraphics[width=0.48\textwidth]{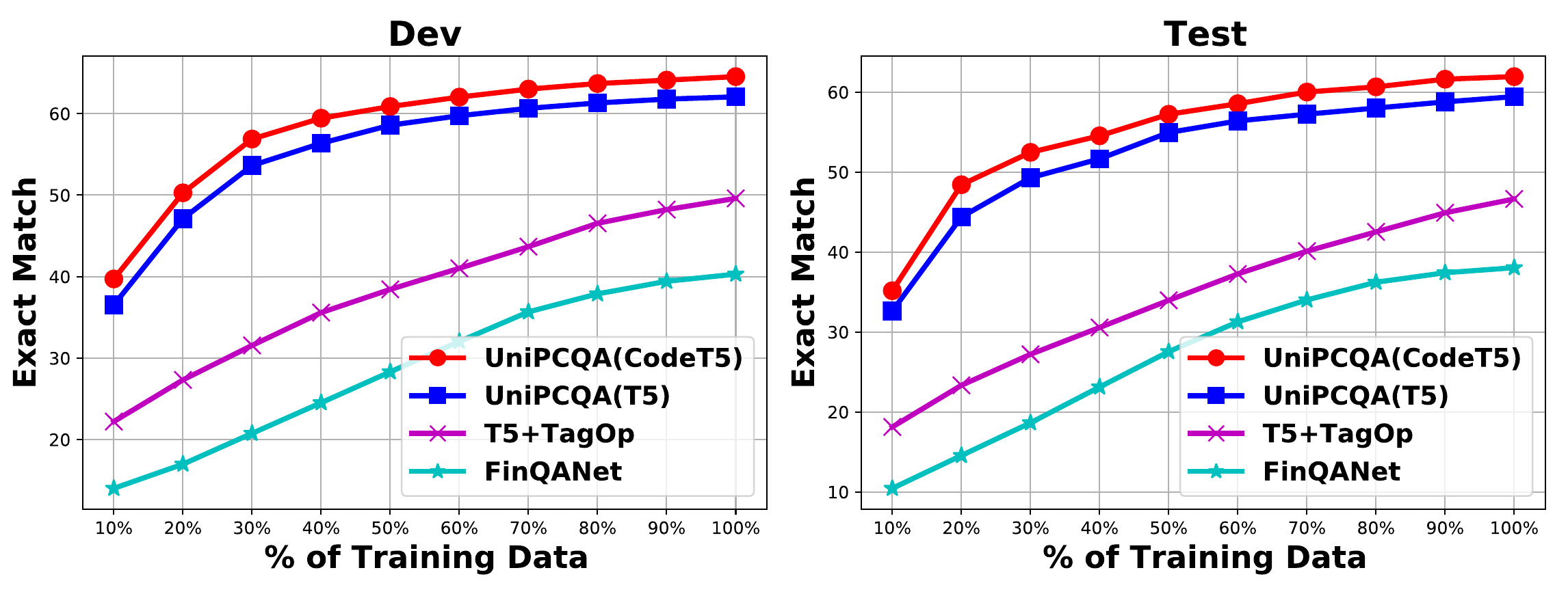}
\caption{Performance w.r.t different \% of training data.}
\label{low-resource}
\end{figure}

\begin{table}
    \centering
    \begin{adjustbox}{max width=0.48\textwidth}
    \begin{tabular}{lrrrr}
    \toprule
         & Table & Text & Table-text & Total \\
         \midrule
       Span  & \textbf{61.3}/57.1 & \textbf{54.8}/50.8 & 70.8/\textbf{76.6} & \textbf{60.6}/59.2\\
       Spans & 61.4/\textbf{67.5} & \textbf{28.6}/23.8 & \textbf{77.4}/72.6 & \textbf{66.2}/65.7\\
       Counting & \textbf{45.5}/36.4 & -&\textbf{44.8}/41.4 & \textbf{45.0}/40.0\\
       Arithmetic & 53.5/\textbf{58.7} & \textbf{27.3}/18.2 & 63.7/\textbf{67.4} &56.1/\textbf{60.7} \\
       Question & 55.6/\textbf{62.2}&\textbf{72.7}/63.6&64.8/\textbf{72.5}& 61.5/\textbf{65.9}\\
       Total & 56.0/\textbf{59.5} &\textbf{54.6}/50.0&67.4/\textbf{70.6}&59.4/\textbf{61.9}\\
       \bottomrule
    \end{tabular}
    \end{adjustbox}
    \caption{Performance comparisons of UniPCQA intialized with T5/CodeT5,  w.r.t different answer types and sources. The results are EM scores on test set.}
    \label{tab:type_source}
\vspace{-0.3cm}
\end{table}

\subsubsection{Detailed Analyses}
\noindent\textbf{Low-resource Evaluation} ~Due to the high expenses in annotations, data is one of the largest bottlenecks for financial QA. 
We investigate how UniPCQA performs w.r.t different number of training data, by splitting 10\% to 100\% of training data for evaluation. 
As shown in Fig.~\ref{low-resource}, compared with FinQANet and T5+\textsc{TagOp}, UniPCQA can better transfer the knowledge from PLMs to achieve a much better performance in low-resource settings, especially from CodeT5.

\noindent\textbf{Answer Type and Source Analysis}
~We further compare the performance of UniPCQA w.r.t answer types and sources. As shown in Table~\ref{tab:type_source}, it can be observed that UniPCQA initialized with CodeT5 or T5 performs differently in terms of answer types and sources. CodeT5 performs much better than T5 on arithmetic questions (60.7 vs. 56.1), which indicates the effectiveness of reformulating the numerical reasoning process as code generation. This leads to better performance in the overall evaluation, since the majority in PACIFIC are arithmetic questions. Conversely,  T5 has better performance on textual data as well as questions relying on span extraction, \textit{e.g.}, Span and Spans. 

\noindent\textbf{Case Study for Consensus Voting} ~Table~\ref{tab:xval} illustrates two examples where the Consensus Voting strategy remedies the mistakes made by greedy decode. In the first example, greedy decode generates a wrong formula, while Consensus Voting derives the correct answer by taking the plurality vote of the results from diverse numerical calculations. 
In the second example, the question is ambiguous as the period is not specified for the percentage change value. Greedy decode makes the wrong prediction on clarification needs, which affects the final answer. However, based on the plurality vote, most sampled outputs in Consensus Voting decide to ask a clarifying question, instead of directly calculating the percentage change value in a random period. 
More details about the case study can be found in Appendix~\ref{app:xval-example}.

\begin{table}
    \centering
    \begin{adjustbox}{max width=0.48\textwidth}
    \begin{tabular}{l|cp{1.2cm}p{7.5cm}}
    \toprule
        Question 1& 
        \multicolumn{3}{l}{What is the average annual amount of it?} \\
        \midrule
        Answer & \multicolumn{3}{l}{2}\\
         \midrule
         & \# & Resp. & Sampled Outputs \\
         \midrule
     Greedy & - & 1.99 & [clari.] False [resp.] (1.06+0.91+\underline{4.01})/3\\ 
         \midrule
     \multirow{2}{*}{CV 1} & \multirow{2}{*}{24} & \multirow{2}{*}{2} & [clari.] False [resp.] (1.06+0.91+4.04)/3 \\
     &&&[clari.] False [resp.] (1.06+4.04+0.91)/3 \\
         \midrule
     CV 2 & 12 & 1.99 & [clari.] False [resp.] (1.06+0.91+\underline{4.01})/3\\
     \midrule
     CV 3 & 4 & 3 & [clari.] False [resp.] (1.06+0.91+4.04)/\underline{2}\\
     \midrule
     \midrule 
        Question 2 & \multicolumn{3}{l}{What is the change in its amount as a percentage?} \\
        \midrule
        Answer &  \multicolumn{3}{l}{Which period are you asking about?}\\
         \midrule
         & \# & Resp. & Sampled Outputs \\
         \midrule
     Greedy  & - & 0.0 & [clari.] \underline{False} [resp.] (576523-576523)/576523\\
         \midrule
     CV 1& 22 & \multicolumn{2}{l}{[clari.] True  [resp.] ['Which period are you asking about?']}\\
         \midrule
     CV 2 & 10 & 0.0 & [clari.] \underline{False} [resp.] (576523-576523)/576523\\ 
         \midrule
     CV 3 & 4 & 7.18 & [clari.] \underline{False} [resp.] (576523-537891)/537891\\
         \midrule
     CV 4 & 2 & -1.8 & [clari.] \underline{False} [resp.] (566523-576891)/576523 \\
    \bottomrule
    \end{tabular}
    \end{adjustbox}
    \caption{Case study for Consensus Voting (CV). The \underline{underlined} content denotes the mistake in the decoded output.}
    \label{tab:xval}
\vspace{-0.4cm}
\end{table}

\subsubsection{Error Analysis}\label{sec:error}
In order to investigate the typical failure cases in UniPCQA, we randomly sample 100 error cases for analysis.  As shown in Table~\ref{tab:error}, we categorize these failure cases into the following six groups: 
\begin{itemize}[leftmargin=*]
    \item \textit{Wrong Evidence} (34\%): The model extracts wrong supporting evidences from the context. 
    \item \textit{Wrong Clarification Need Prediction} (18\%): The model makes a wrong prediction on whether the user query requires clarification. More specific, 11\% of all the failure cases are predicted to be unnecessary for clarification, while they are ambiguous in fact. And 7\% of them vice versa. 
    \item \textit{Wrong Derivation} (13\%): Although the model extracts all the necessary supporting evidences, the model fails to compute the answer with a correct derivation, \textit{e.g.}, wrong formula or order. 
    \item \textit{Missing Evidence} (12\%): Although the extracted evidences are correct, the model fails to extract all the required evidences from the context. 
    \item \textit{Wrong Clarification Question} (7\%): The model generates a wrong clarification question that fails to clarify the ambiguity of the user query. 
    \item \textit{Other Errors} (18\%): There are several other errors that are relatively acceptable, such as the scale error, missing symbols or missing punctuation marks. 
\end{itemize}
Compared with the error analysis of the \textsc{TagOp} model in the TAT-QA dataset~\cite{tat-qa}, it is worth noting that the percentage of errors that related to span extraction largely decreases from 84\% to 46\%. However, there are about 25\% and 13\% of errors that are related to the clarification question generation task and the numerical calculation, respectively.

\begin{table}[]
\fontsize{8.5}{10}\selectfont
    \centering
    \begin{adjustbox}{max width=0.5\textwidth}
    \begin{tabular}{p{1.5cm}|p{5.2cm}}
    \toprule
       Wrong & Q: What was the change in its amount in \\
       Evidence &  2019 from 2018?\\
       (34\%) &  G: 2.1 - 1.8 \\
       & P: 2.1 - \underline{1.3}\\
       \midrule
       Wrong & Q: In which year were the PSP payments\\
       Clarification &  larger?\\
       Need & G: What kind of PSP payments are you\\     
       Prediction &  asking about?\\
       (18\%) & P: 2019\\
       \midrule
       Wrong & Q: How about their average salary? \\
       Calculation & G: (1,000,000 + 650,000 + 440,000) / 3  \\
       (13\%) & P: (1,000,000 + 650,000 + 440,000) / \underline{2}\\
       \midrule
       Missing &Q: What is the total stock-based compen-\\
       Evidence &sation expense and unrecognized stock-\\
       (12\%) & based compensation expense in 2019? \\
       & G: 3,711 + \underline{4,801} + 1,882 \\
       & P: 3,711 + 1,882\\
       \midrule
       Wrong & Q: What is the total long-term debt due?\\
       Clarification & G: Which period of payments due are you\\
       Question & asking about?\\
       (7\%) &    P: Which year are you asking about?\\
       \midrule
       Other & Q: What was the cash and cash equivalents\\
       Errors & in 2018? \\
       (18\%) & G: \$148,502\\
        &   P: 148,502\\
       \bottomrule
    \end{tabular}
    \end{adjustbox}
    \caption{Error Analysis (G: Ground-truth, P: Prediction).}
    \label{tab:error}
    \vspace{-0.4cm}
\end{table}

\section{Conclusions}
In this paper, we present a new dataset, PACIFIC, for proactive conversational question answering over a hybrid context of tables and text.
Accordingly, we define the problem of Proactive Conversational Question Answering that combines clarification question generation and conversational question answering. 
In addition, we reformulate the numerical reasoning process as code generation and recast all sub-tasks in PCQA into a Seq2Seq problem solved by a unified model, UniPCQA. 
Extensive experiments show that the PACIFIC dataset is very challenging and demonstrate the need to build models that can handle hybrid input and output formats as well as diverse numerical reasoning.

\section*{Ethical Considerations}
The PACIFIC dataset was built from the TAT-QA dataset, which is publicly available. The authors of the TAT-QA dataset paper have allowed us to utilize the dataset for further construction. 
We will provide open access to our dataset and code for future studies via \url{https://github.com/dengyang17/PACIFIC/}.  

\section*{Limitations}
In this section, we analyze the limitations from the perspectives of both the constructed dataset and the proposed method.

\subsection*{Limitations of PACIFIC Dataset} 

Since PACIFIC is the first CIS dataset in finance domain as well as the first proactive CQA dataset, there are inevitably some limitations and room for further improvement. 
\begin{itemize}[leftmargin=*]
    \item \textbf{Numerical Reasoning}. Similar to other popular NLP datasets that require numerical reasoning, such as DROP~\cite{drop} and FinQA~\cite{finqa}, the questions in PACIFIC only require some basic numerical calculations, including arithmetic operations, counting, and comparison. In the future, with the advance in the model capability of numerical reasoning, it would be better to add questions that require more complicated numerical calculations. 
    \item \textbf{Clarification Question}. In the clarification turn, PACIFIC only provides the clarification question. In some cases, it is beneficial to further provide the candidate options for better clarifying the uncertainty~\cite{emnlp19-clari,mimic}. 
    Besides, in the data creation process, we construct ambiguous questions that contain only one missing information for guaranteeing the objectivity of the clarification question annotations. However, it is also worth studying the situation where there are multiple missing information for clarification. 
    \item \textbf{Multimodality}. Although the PACIFIC dataset is based on a hybrid context of tables and text, there are more diverse information in the real-world financial documents with different modalities, such as images, charts, etc. It is necessary to consider more comprehensive QA or CQA datasets and problem settings for real-world applications in finance domain. 
\end{itemize}

\subsection*{Limitations of UniPCQA}
The error analysis in Section~\ref{sec:error} reveals some limitations in the proposed method. 
Currently, the capability of numerical reasoning in UniPCQA relies on the pre-trained language models. 
In the future, we would like to investigate post-training strategies to transfer task-adaptive or domain-specific knowledge from other post-training tasks for further improving this capability. 
In addition, due to the heterogeneous input and output content, it would also be beneficial to investigate more robust prompt-based learning approaches for better learning the relationships among different types of information.

\bibliography{emnlp2022}
\bibliographystyle{acl_natbib}

%\clearpage

\appendix

\begin{figure*}
\centering
\includegraphics[width=\textwidth]{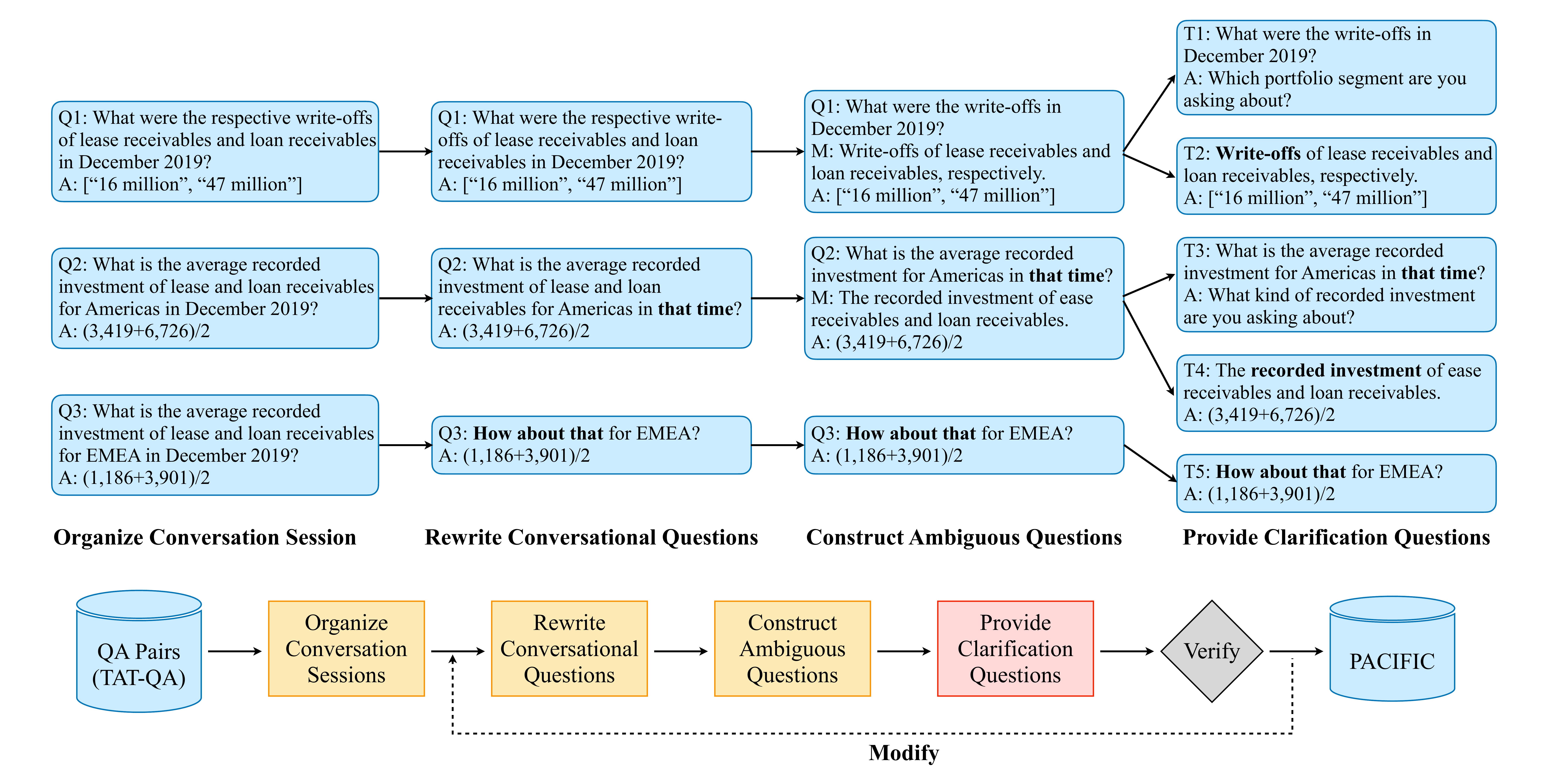}
\caption{Overall Pipeline for PACIFIC Creation with Examples.}
\label{pipeline}
\end{figure*}

\section*{Appendix}
\section{Pipeline of Dataset Creation}\label{app:dataset}
Fig.~\ref{pipeline} presents the illustration of overall pipeline for the PACIFIC dataset creation with examples. 
In financial question answering~\cite{tat-qa,finqa}, the user query is supposed to be informative and complicated with multiple constraints. 
Therefore, it is inappropriate to adopt the traditional way of decomposing a complex single-turn question into multiple conversational questions with limited information for constructing a financial conversational question answering dataset.  

To this end, we employ a different pipeline to create the PACIFIC dataset. 
As described in Section~\ref{sec:dataset}, there are totally four steps for the creation of the PACIFIC dataset, including (1) Organize Conversation Sessions\footnote{Entities in the question will be automatically highlighted for the convenience of annotators, through lexical matching with the nouns in paragraphs and tables.}, (2) Rewrite Conversational Questions, (3) Construct Ambiguous Questions\footnote{Only one entity in the original question is randomly chosen to be omitted for the annotators to construct the ambiguous question.}, and (4) Provide Clarification Questions\footnote{Similar to \newcite{www2020-clari}, we provide several templates for the annotators to provide clarification questions using the omitted entity. This guarantees the objectivity of the clarification question annotations}. 
This annotation pipeline not only increases efficiency in the dataset construction, but also guarantees the quality and preserves the difficulty of the dataset with expert-annotated answers and informative user queries for financial CQA.

\section{Implementation Details}
\label{app:implement}
The pre-trained weights of T5 and CodeT5 are initialized using HuggingFace\footnote{https://huggingface.co}. 
We use the same hyper-parameter settings for different initialization. The learning rate and the weight decay rate are set to be 5e-5 and 0.01, respectively. The max source sequence length and the max target sequence length are 1280 and 128, respectively. 
We train the model up to 15 epochs with mini-batch size of 4, and select the best checkpoints based on the EM score on the validation set. 
We train the model on three NVIDIA Tesla V100 GPUs with 32GB RAM.

For a fair comparison, all the PLM-based baselines adopt the version of PLMs with a similar size of model parameters as T5$_\text{base}$ (220M) and CodeT5$_\text{base}$ (220M). For example, BERT, RoBERTa, and GPT-2 based methods adopt BERT$_\text{large}$ (340M), RoBERTa$_\text{large}$ (355M), and GPT-2$_\text{medium}$ (345M), respectively. 

\section{Compared Baselines}
\label{app:baseline}
Following \newcite{tat-qa}, we simply flatten the table into a sequence by row followed by tokens from the paragraphs for all the baselines. 

\subsection{Clarification Need Prediction}
We fine-tune the vanilla BERT$_\text{large}$~\cite{bert} and RoBERTa$_\text{large}$ \cite{roberta} based classifiers for the CNP task. 

\subsection{Clarification Question Generation}
We compare to the following CQG baselines:
\begin{itemize}[leftmargin=*]
    \item \textbf{Template-based Question Generation (TB)}: A template-based approach~\cite{www2020-clari} to generating clarifying questions produces a question by simply filling a slot in a pre-defined question, i.e., ``What kind of \_ are you asking about?''. And we adopt a fine-tuned BERT-based span extraction model to extract the slot value from the document. 
    \item \textbf{CopyTrans.}~\cite{cikm21-cqg} adopts the Transfomer-based encoder-decoder with the copy mechanism for the CQG task. 
    \item \textbf{Q-GPT}~\cite{ictir21-qgpt} fine-tunes GPT-2~\cite{gpt2} to generate clarifying questions. 
\end{itemize}

\subsection{Query Rewriting}
We adopt the following QR baselines for evaluation:
\begin{itemize}[leftmargin=*]
    \item \textbf{Original}: Use the original conversational question at the current conversation turn without query rewriting, which is often regarded as the lower bound for the pipeline-based CQA evaluation. 

    \item \textbf{Trans.++}: A Transformer-based QR model \cite{wsdm21-query-cqa} initialized with the weights of pre-trained GPT-2 model~\cite{gpt2}. 

    \item \textbf{T5}: Following~\cite{cqr-t5,acl21-coqa-dependency}, we adopt a T5-based sequence generator~\cite{t5} as a baseline QR model. 
\end{itemize}

\subsection{Conversational Question Answering}
We adopt the following QA and CQA baselines for evaluation:
\begin{itemize}[leftmargin=*]
    \item \textbf{NumNet+ V2}\footnote{https://github.com/llamazing/numnet\_plus}: A numerical QA model utilizes a numerically-aware graph neural network to consider the comparing information and performs numerical reasoning over texts~\cite{emnlp19-numnet}.
    \item \textbf{\textsc{TagOp}}\footnote{https://github.com/NExTplusplus/TAT-QA}: A RoBERTa-based QA model adopts sequence tagging to extract information and applies numerical reasoning over tables and texts with a set of aggregation operators~\cite{tat-qa}. 
    \item \textbf{BERT+HAE}\footnote{https://github.com/prdwb/bert\_hae}: A BERT-based CQA model adds the history answer embeddings (HAE) to the BERT's word embeddings~\cite{sigir19-coqa-history}. Since there is no numerical reasoning module in this method, the output only contains the extracted spans from the documents. 
\end{itemize}

\subsection{Proactive Conversational Question Answering}
We adapt the following methods for the evaluation of the overall PCQA problem:
\begin{itemize}[leftmargin=*]
    \item \textbf{DialoGPT} ~Following \citet{hybridialogue}, we fine-tuned a pre-trained DialoGPT model~\cite{dialogpt} for dialogue response generation. Similar to BERT+HAE, the target sequence only contains the required spans from the document without numerical calculations. 

    \item \textbf{FinQANet}\footnote{https://github.com/czyssrs/finqa}~\cite{finqa} ~A retriever using BERT first retrieves the supporting facts from the document, then a generator combining RoBERTa and LSTM generates the response, which can be either a schema-based program for CQA or a natural language question for CQG. 

    \item \textbf{T5+\textsc{TagOp}} A pipeline-based method first uses T5~\cite{t5} for the sub-tasks of CNP and CQG. If it is not required for clarification, \textsc{TagOp}~\cite{tat-qa} is adopted to produce the answer as an end-to-end CQA method.  
\end{itemize}
Note that for two end-to-end methods, including DialoGPT and FinQANet, there is no sub-task of CNP, while the whole PCQA problem can be regarded as the response generation problem in dialogue systems. 

\begin{table}
\fontsize{8}{9.5}\selectfont
    \centering
    \begin{tabular}{lccccc}
    \toprule
       \multirow{2}{*}{QR Model} & \multirow{2}{*}{Train Set} & \multicolumn{2}{c}{Dev} & \multicolumn{2}{c}{Test}\\
       \cmidrule(lr){3-4}\cmidrule(lr){5-6}
        & & ROUGE & EM  & ROUGE & EM \\
        \midrule
        Original & - & 68.8& 43.3&69.9&43.6\\
        Trans.++ & QReCC &78.2&28.9&77.4&30.2\\
        T5 & QReCC & 79.5&28.7&80.2&29.7\\
        Trans.++ & PACIFIC &93.6&76.5&92.9&76.4\\
        T5 & PACIFIC & 94.9&80.1&94.8&79.7\\
        \bottomrule
    \end{tabular}
    \caption{Results on Query Rewriting.}
    \label{tab:cqr}
\end{table}

\begin{figure}
\centering
\includegraphics[width=0.4\textwidth]{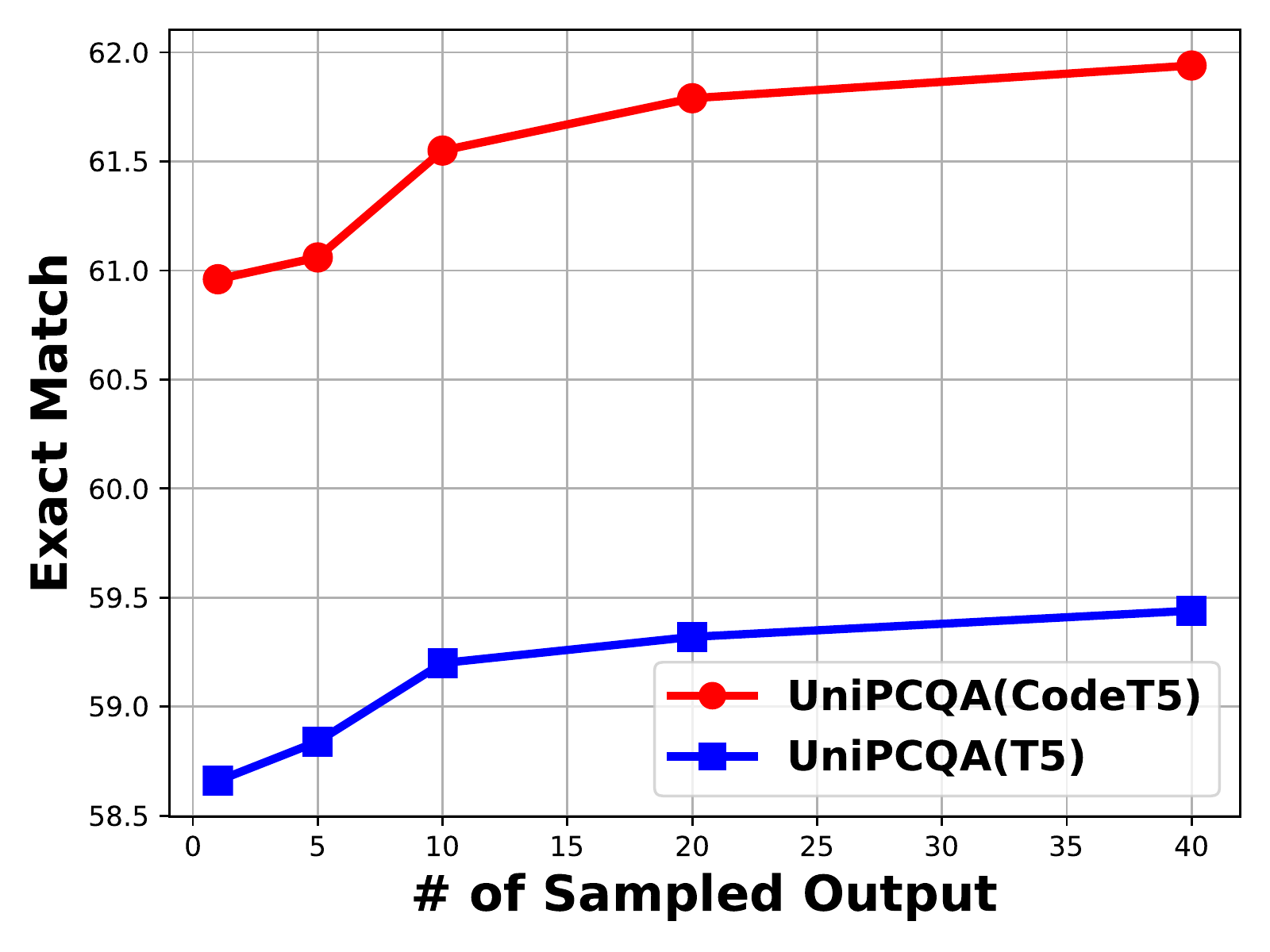}
\caption{Performance w.r.t different number of sampled output.}
\label{num_sample}
\end{figure}

\section{Evaluation on Query Rewriting}\label{app:qr}
Following previous studies~\cite{wsdm21-query-cqa}, we adopt ROUGE-1 (Recall) and EM for the evaluation of QR models. 

The performance of each Query Rewriting (QR) method is presented in Table~\ref{tab:cqr}. 
Due to the substantial difference between financial CQA and general CQA, \textit{i.e.}, PACIFIC and QReCC, the QR models trained on QReCC perform poorly in PACIFIC.

\section{Effect of Sampling Number}
\label{app:sample_num}
Fig.~\ref{num_sample} shows the performance of Consensus Voting in terms of different number of sampled outputs from the decoder, ranging from [1, 5, 10, 20, 40]. 
Due to the restriction of experimental environment, the maximum number of sample outputs is set to be 40. 
Experimental results show that a higher number of sampled outputs generally leads to a better performance for both T5 and CodeT5-based UniPCQA models, which indicates the effectiveness of the plurality voting in the Consensus Voting strategy for alleviating the error propagation issue.

\begin{figure}
\centering
\includegraphics[width=0.5\textwidth]{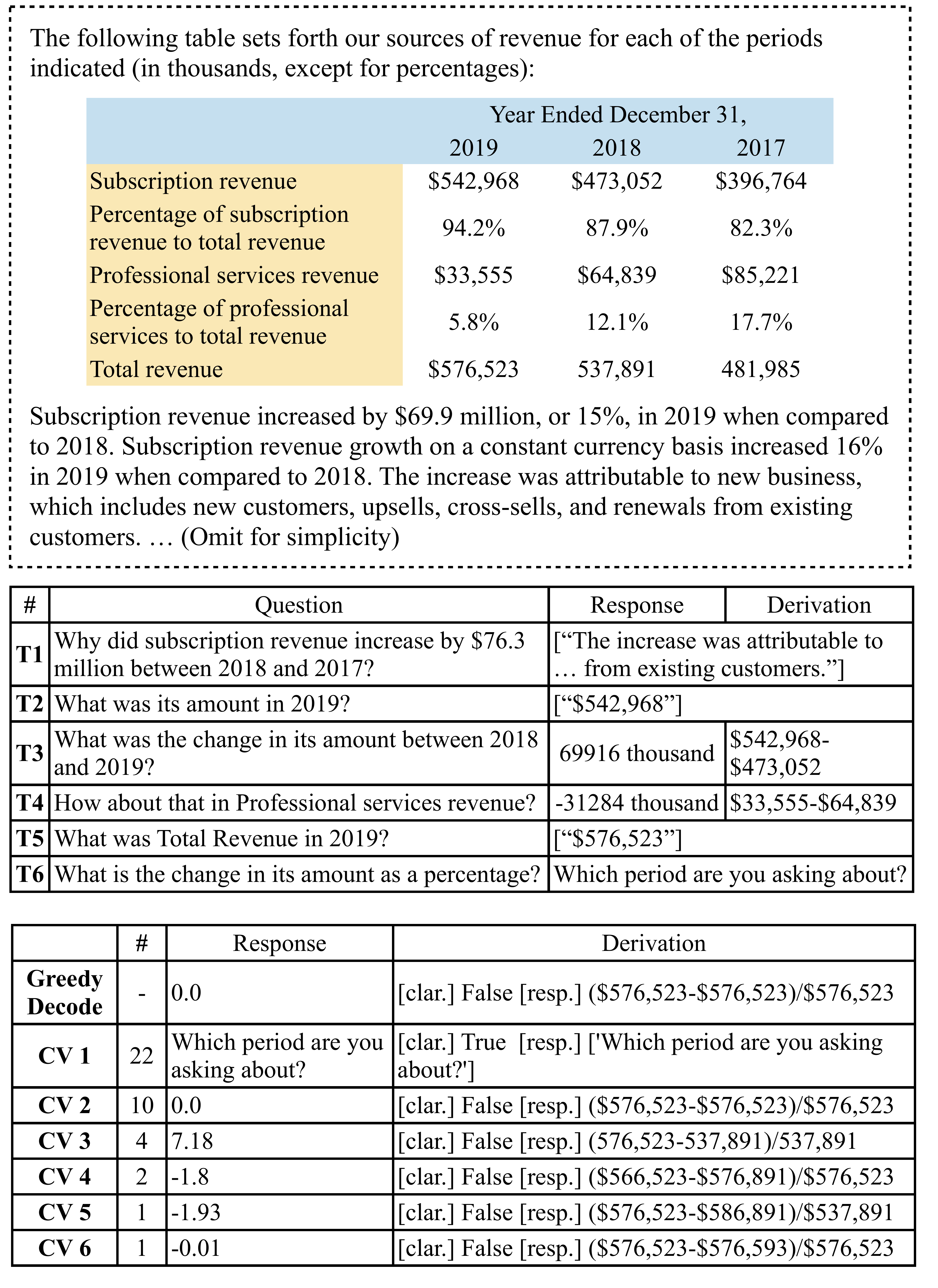}
\caption{Detailed Case Study for Consensus Voting (CV).}
\label{case1}
\end{figure}

\section{Detailed Case Study}\label{app:xval-example}
Fig.~\ref{case1} presents the details of the case study in Table~\ref{tab:xval}, including the grounded document and the conversation history. 
At the current turn \textbf{T6}, the user query is ``What is the change in its amount as a percentage?", where ``it" refers to ``\textit{Total Revenue}" at the previous turn \textbf{T5}. 
As shown in the tabular context, the amount of ``\textit{Total Revenue}" is recorded in three years, from 2017 to 2019. 
Due to the uncertainty of the concerned period, the user query, ``What is the change in its amount as a percentage?", is ambiguous under this context. 

In the inference, the decoder will be confused about which figures are supposed to be extracted from the grounded document. 
We can observe that Greedy Decode generates a wrong derivation, which may answer the query ``What is the change in its amount as a percentage from 2019 to 2019?".
Similarly, for CV 3, the generated derivation is supposed to answer the query ``What is the change in its amount as a percentage from 2018 to 2019?". 
However, at this conversation turn, the system is not aware of the specific period that the user is asking.   
Therefore, all these derivations with a random period are incorrect. 
Overall, with the plurality voting, Consensus Voting effectively alleviates this kind of issue, since it would be difficult for the sampled derivation outputs to make a consensus for an ambiguous question. 

\end{document}